\documentclass[%
 aip,
 amsmath,amssymb,
 reprint,%
]{revtex4-1}

\usepackage{graphicx}% Include figure files
\usepackage{dcolumn}% Align table columns on decimal point
\usepackage{bm}% bold math
\usepackage{amsmath,amssymb,amsthm}

\usepackage[mathscr]{euscript}
\DeclareSymbolFont{rsfs}{U}{rsfs}{m}{n}
\DeclareSymbolFontAlphabet{\mathscrsfs}{rsfs}
\usepackage[utf8]{inputenc}
\usepackage[T1]{fontenc}
\usepackage{mathptmx}
\usepackage{etoolbox}
\usepackage{booktabs} % Add this in the preamble

\makeatletter
\def\@email#1#2{%
 \endgroup
 \patchcmd{\titleblock@produce}
  {\frontmatter@RRAPformat}
  {\frontmatter@RRAPformat{\produce@RRAP{*#1\href{mailto:#2}{#2}}}\frontmatter@RRAPformat}
  {}{}
}%
\makeatother
\begin{document}

\preprint{AIP/123-QED}

\title[%small sttle
]{Constrained Bayesian Optimization Using a Lagrange Multiplier Applied to Power Transistor Design}
% Force line breaks with \\
\author{Ping-Ju Chuang}
\affiliation{%
Department of Materials Science and Engineering, The University of Texas at Dallas, Richardson, TX 75080 USA}%

\author{Ali Saadat}
\affiliation{%
Kilby Labs, Texas Instruments Inc., Santa Clara, CA 95051 USA}%

\author{Sara Ghazvini}
\affiliation{%
Department of Materials Science and Engineering, The University of Texas at Dallas, Richardson, TX 75080 USA}%

\author{Hal Edwards}%
 \affiliation{ 
Analog Technology Development, Texas Instruments Inc., Dallas, TX 
75243 USA}%

\author{William G. Vandenberghe}
 \email{william.vandenberghe@utdallas.edu.}
\affiliation{%
Department of Materials Science and Engineering, The University of Texas at Dallas, Richardson, TX 75080 USA}%

% \date{\today}% It is always \today, today,
%              %  but any date may be explicitly specified

\begin{abstract}
We propose a novel constrained Bayesian Optimization (BO) algorithm optimizing the design process of Laterally-Diffused Metal-Oxide-Semiconductor (LDMOS) transistors while realizing a target Breakdown Voltage ($BV$). We convert the constrained BO problem into a conventional BO problem using a Lagrange multiplier. Instead of directly optimizing the traditional Figure-of-Merit (FOM), we set the Lagrangian as the objective function of BO. This adaptive objective function with a changeable Lagrange multiplier can address constrained BO problems which have constraints that require costly evaluations, without the need for additional surrogate models to approximate constraints. Our algorithm enables a device designer to set the target $BV$ in the design space, and obtain a device that satisfies the optimized FOM and the target $BV$ constraint automatically. Utilizing this algorithm, we have also explored the physical limits of the FOM for our devices in 30 - 50 V range within the defined design space.
\end{abstract}

\maketitle

\section{\label{sec:level1}Introduction}

Efficient power electronics are an essential component of a society powered by renewable energy. Power semiconductor devices have been extensively adapted in all applications that require electrical power. Different materials such as silicon\cite{baliga2010fundamentals}, Silicon carbide (SiC)\cite{kimoto2014fundamentals, she2017review}, and Gallium nitride (GaN)\cite{meneghini2021gan, 7862945}, along with various device structures like power diodes, planar/trench Metal-Oxide-Semiconductor Field-Effect Transistors (MOSFETs)\cite{williams2017trench, williams2017trench2}, and Insulated-Gate Bipolar Transistors (IGBTs)\cite{iwamuro2017igbt, khanna2004insulated}, are utilized in the fabrication of a wide range of analog microchips, catering to the power needs of diverse power management applications. The choice of material and device structure depends on the operating voltage requirements and specific performance characteristics desired for each application. These technologies enable the provision of electrical power to various appliances and systems. Among all power semiconductor devices, silicon-based Laterally-Diffused Metal-Oxide-Semiconductor (LDMOS) transistors are the most popular device since they can be seamlessly integrated into Integrated Circuit (IC) technology\cite{chen2021study, chou20120, jin2017best, huang20140, cha20160, 9761987, saadat2020simulation}, bringing economic benefits and being able to bridge an extensive voltage range.

The Figure-of-Merit (FOM) = $BV^2/R_{\rm sp(on)}$ is an important metric employed to assess the quality of power devices~\cite{FOMs}. This criterion reflects the maximum achievable power density in a power device which is crafted by specific materials. As the LDMOS transistor is integrated within the silicon process and widely used to design various devices working at different operating voltages, the design challenge for LDMOS transistors has shifted to minimize the $R_\mathrm{sp(on)}$ at specific voltages to approach the material's physical limits and reduce the cost simultaneously. This principle of device design, known as the ''power density scaling`` law has been consistently applied in the field of power device design. Following this law, consequently, the ongoing challenge in the industry lies in designing and optimizing different LDMOS transistors with disparate $BV$ values within the same semiconductor process.

In Moore's Law transistors, all parameters such as the operation voltage are chosen for accurate and efficient communication, aiming to reduce cost-per-transistor for every new technology node~\cite{roadmap}. In power semiconductor technology, however, microchips process power flows across different voltage domains specific to each application. For instance, a laptop power converter must efficiently convert direct current (DC) power from a 12 V battery to drive a 1 V microprocessor. Similarly, automotive applications may require transient operation at 40 V or higher due to fault conditions which exist in the automotive environment. Power transistors must support a wide voltage range, and their breakdown voltage ($BV$) needs to be substantially higher than the operating voltage to ensure reliability throughout their operation lifetime. For example, the laptop 12 V to 1 V DC power converter mentioned above may require $BV$ > 30 V, whereas an automotive microchip that survives fault conditions at 40 V might need a device with $BV$ > 55 V. Thus, the optimization of power transistors involves selecting a limited range of $BV$ values, considering specific application requirements.

Bayesian optimization (BO) is a data-driven algorithm that efficiently determines the global extremum within a defined space while minimizing the number of evaluations required~\cite{garnett_bayesoptbook_2023, frazier2018tutorial}. In recent years, BO has gained significant popularity as a replacement for grid search or random search in optimizing the architecture of different machine learning models\cite{bergstra2012random, snoek2012practical, bergstra2011algorithms}. Moreover, BO is being increasingly adopted in various research domains to accelerate the optimization process and enhance the development of their respective fields such as chemical synthesis\cite{shields2021bayesian}, process of solar cells\cite{xu2023bayesian}, optimization of lithium-ion batteries\cite{gaonkar2022multi}, and drug discovery\cite{guan2022class}. Ironically, in more mature fields like the semiconductor industry, due to its earlier development compared to the popularity of BO, there is relatively less use of BO or other artificial intelligence assisted tools. We have previously integrated different optimization algorithms with Technology computer-aided design (TCAD) simulations to automate the optimization of semiconductor devices~\cite{mypaper}. Among all these optimization algorithms, BO is the most data efficient method to integrate with TCAD device simulations.

TCAD simulation is an indispensable tool in semiconductor device manufacturing. Due to the high cost associated with semiconductor device fabrication, performing TCAD simulations and designing desired device characteristics is a crucial part of device fabrication. Skilled engineers perform a large number of TCAD simulations, often guided by design of experiments (DOE) approaches like factorial designs, response surface designs, or Taguchi designs~\cite{chelladurai2021optimization, 4736103}. The combination of DOE and TCAD simulations requires significant manual intervention and human involvement in the process. By integrating BO with TCAD simulations, we can perform automated and globally optimized designs, accelerating the development process. However, while optimizing device design, constraints must be respected. e.g., when designing a power device, $BV$ needs to be sufficiently high. Checking whether a constraint ($BV$) is satisfied can be the most computationally expensive part of simulations. Previous constrained BO relies on different surrogate models to approximate both the output parameter ($BV$) and the objective function (FOM) \cite{gardner2014bayesian}. However, this approach requires using multiple surrogate models, which increase the time and resources required for evaluations to solve such constrained BO problems.

In this paper, we propose a constrained BO process using a Lagrange multiplier. We apply the methodology to the optimization of an LDMOS transistors enabling the optimization of LDMOS transistors with a given breakdown voltage. We first show that the unconstrained BO yields a device with a breakdown voltage of 31~{\rm V} but reveals no information about devices with $BV$ of $40~{\rm V}$ or $50~{\rm V}$. Next, we show that we can successfully constrain the $BV$ and find high-FOM devices for $BV=40~{\rm V}$ and $BV=50~{\rm V}$. Our results demonstrate that constrained BO can effectively accomplish the optimization of device designs with specific $BV$ restrictions, and even explore the physical limits for a specific breakdown voltage range without requiring human intervention.

\section{Results}
\begin{figure}
\includegraphics[width=3.5in]{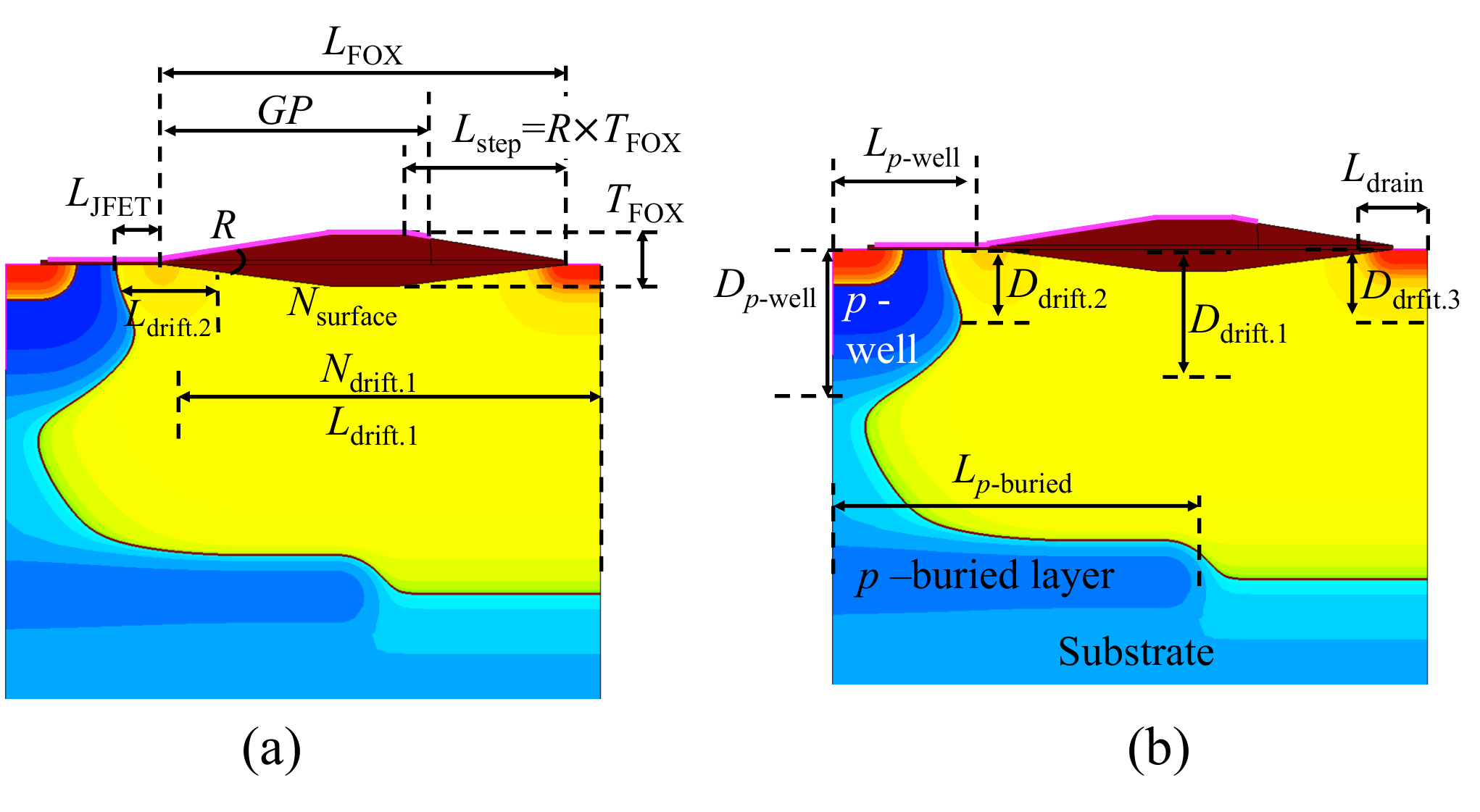}
\caption{\label{fig:device structure} Schematic cross-section of an LDMOS with elongated diamond LOCOS. (a) nine input parameters and (b) ten fixed parameters are labeled. TABLE ~\ref{tab:table1} lists nine input parameters and their bounds.}
\end{figure}

The LDMOS transistors under investigation are illustrated in Fig.~\ref{fig:device structure}. We use the standard Gaussian doping setting in the commercial simulation software to define all the doping profiles in this study~\cite{Sentaurus}. The elongated-diamond-shaped oxide region on the top of the device represents a LOCal-Oxidation Structure (LOCOS). The LOCOS acts as a field-relief oxide and increases the breakdown voltage when designed appropriately. The Junction Field-Effect-Transistor (JFET) region is located on the drift region which is not covered by the LOCOS. The fabrication process of LDMOS transistors with LOCOS can be found on Ref.~\cite{7969235}. The doping concentration in the channel is chosen so that the leakage current remains smaller than $10^{-13}~$A/$\mu$m. The device under consideration has 19 parameters to define the device structure. For the BO, we optimize nine input parameters while keeping ten other input parameters fixed. For detail about constructing LDMOS transistors with LOCOS by 19 parameters in TCAD, we refer to the subsection~\ref{Device structure simulation}.  % 1 paragraph on the device

The FOM = $BV^2/R_{\rm sp(on)}$ is the objective function of our unconstrained BO. $BV$, $R_\mathrm{sp(on)}$, and FOM are illustrated in the methodology section. The first 10 device configurations are chosen randomly followed by 190 or 790 new device structures generated iteratively using the acquisition function which maximizes the expected improvement of the FOM or the Lagrangian. Gaussian process regression is used to emulate the ground truth response surface in our design space. We first show the results of an unconstrained BO, followed by a constrained BO by developing a Lagrangian approach, finally we determine the frontier of the FOM vs $BV$.

%\subsection{Unconstrained BO}

\begin{figure*}
\includegraphics[width=18.2cm]{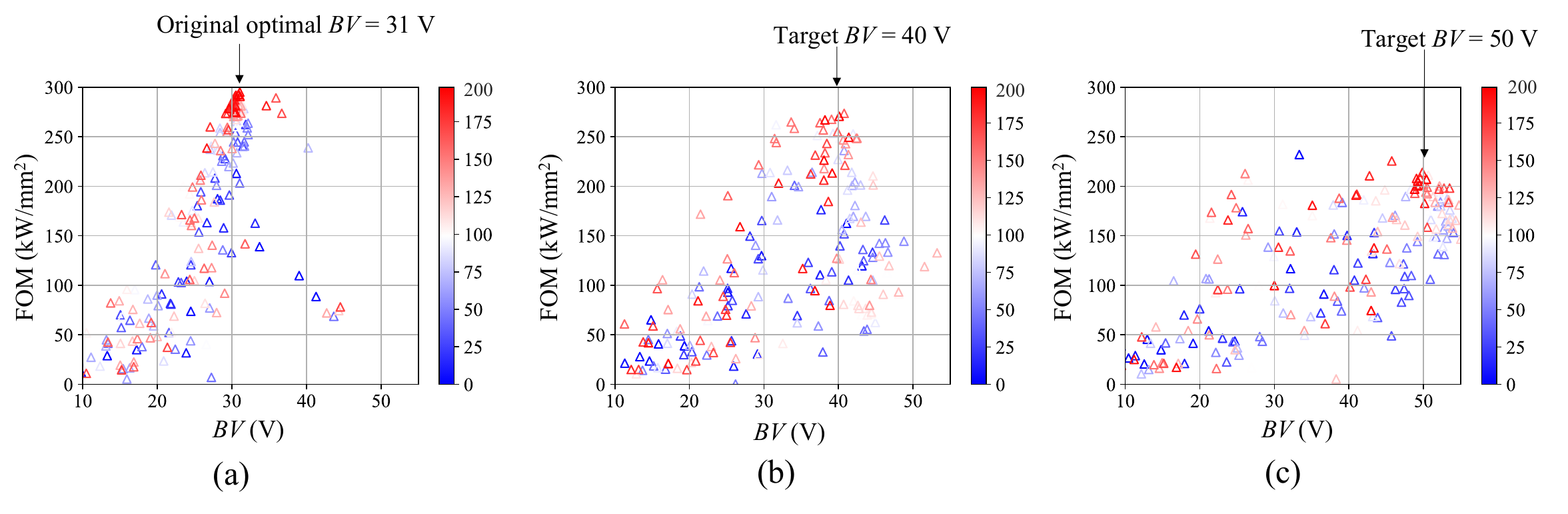}
\caption{\label{fig2} The BO result (a) without any constraint (b) with a constrained target $BV$ = 40 V, (c) constrained $BV$ = 50 V. The color bar indicates the progress of the BO. A deeper red color means the data point is simulated closer to the final step. Fig.~2(b-c) indicate that by performing a constrained optimization, the optimization result changed from the original optimal $BV$ = 31 V to the constrained target  $BV$ = 40 V and 50 V.} % Need to add a device schematic, you can just do 1 colorbar. Also, make the colorbar go from 1 to 200
\end{figure*}

Fig.~\ref{fig2} (a) exhibits the FOM and $BV$ for 200 devices simulated in an unconstrained BO. The color of each marker indicates the order in which the devices are simulated with the darkest blue device simulated first, and the brightest red device simulated last. Most of the devices that are simulated last are found near the device with the highest FOM = 295~$k\mathrm{W/mm}^{2}$. This highest FOM device is found to have $BV$ = 31~V. The vast majority of simulated devices have a $BV$ < 31~V, a few high FOM devices are identified with $BV\approx 35{\rm V}$ while no high-FOM devices with $BV > 40~{\rm V}$ are simulated during the BO. For applications requiring $BV > 40~{\rm V}$, the unconstrained FOM does not provide insight in device design.

%\subsection{Constrained BO}

Fig.~\ref{fig2} (b) and (c) present 200 devices simulated in a constrained optimization with the target $BV$ = 40 and 50 V constraints respectively. For the constrained BO, we use the same approach as for the unconstrained BO except that we optimize a Lagrangian function as detailed in the method section. Fig.~\ref{fig2} (b) has a maximum FOM = 270 $k$W/$\mathrm{mm}^2$ at $BV$ = 41 V while Fig.~\ref{fig2} (c) has identified a FOM = 207 $k$W/$\mathrm{mm}^2$ device with a $BV = 50~{\rm V}$. The device with FOM = 207 $k$W/$\mathrm{mm}^2$ at $BV=50~{\rm V}$ is not the highest FOM device simulated in Fig.~\ref{fig2} (c) but it is the highest FOM device that can realize a $50~{\rm V}$ breakdown voltage. Looking visually at the distribution of the devices in Fig.~\ref{fig2} (a) - (c), by adding the Lagrange term, many more devices with higher $BV$ are simulated.

%\subsection{Determining the frontier}

\begin{figure}
\includegraphics[width=8cm]{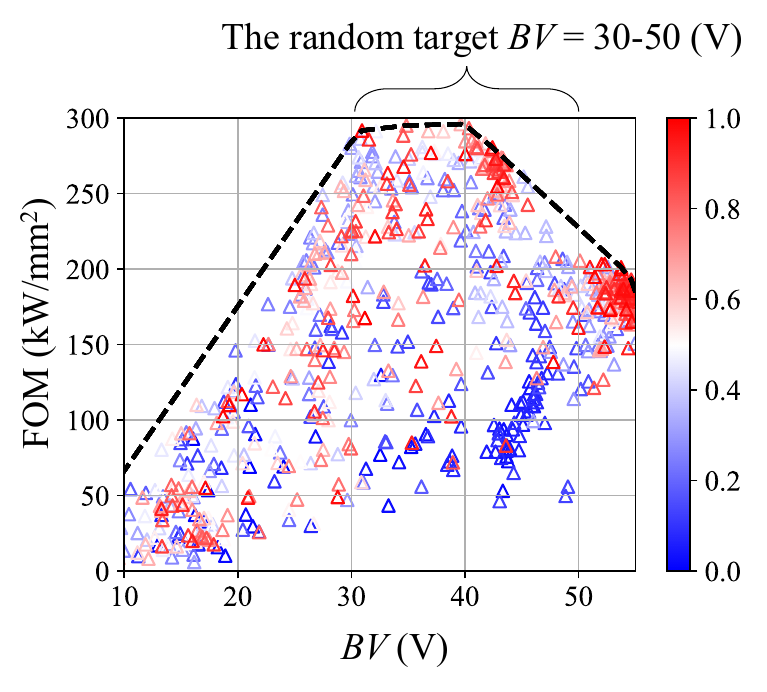}
\caption{\label{fig3} The upper boundary shows the physical limitation of FOM for LDMOS transistors with LOCOS in our design space. We perform 800 simulations and assign target $BV$ randomly between 30 - 50 V in each iteration. }
\end{figure}

Fig.~\ref{fig3} reveals the frontier of FOM vs $BV$. We obtain the device configurations in Fig.~\ref{fig3} by performing a constrained BO where the constraint is changed at every iteration step. In each iteration, a target $BV$ is chosen randomly between 30~{\rm V} and 50~V and the acquisition function chooses a new design based on the Lagrange multiplier associated with the target $BV$ of the current iteration and all previously acquired FOM and $BV$ data. A total of 800 devices are simulated and the upper hull of the obtained results is indicated. Interestingly, the device with the highest FOM = 299 $k$W/$\mathrm{mm}^2$ is found to have a $BV$ = 40 V, exceeding the best device we found from the unconstrained BO with 200 iterations in Fig.~\ref{fig2} (a). Within the design space that we have defined, we find that FOM $\approx$ 300 $k$W/$\mathrm{mm}^2$ devices can be designed with a $BV$ = 30~V to $BV$ = 40~V. Devices with a $BV$ up to 55~V can be designed but the FOM would decrease to $\approx$ 200 $k$W/$\mathrm{mm}^2$.

\begin{figure*}
\includegraphics[width=18cm]{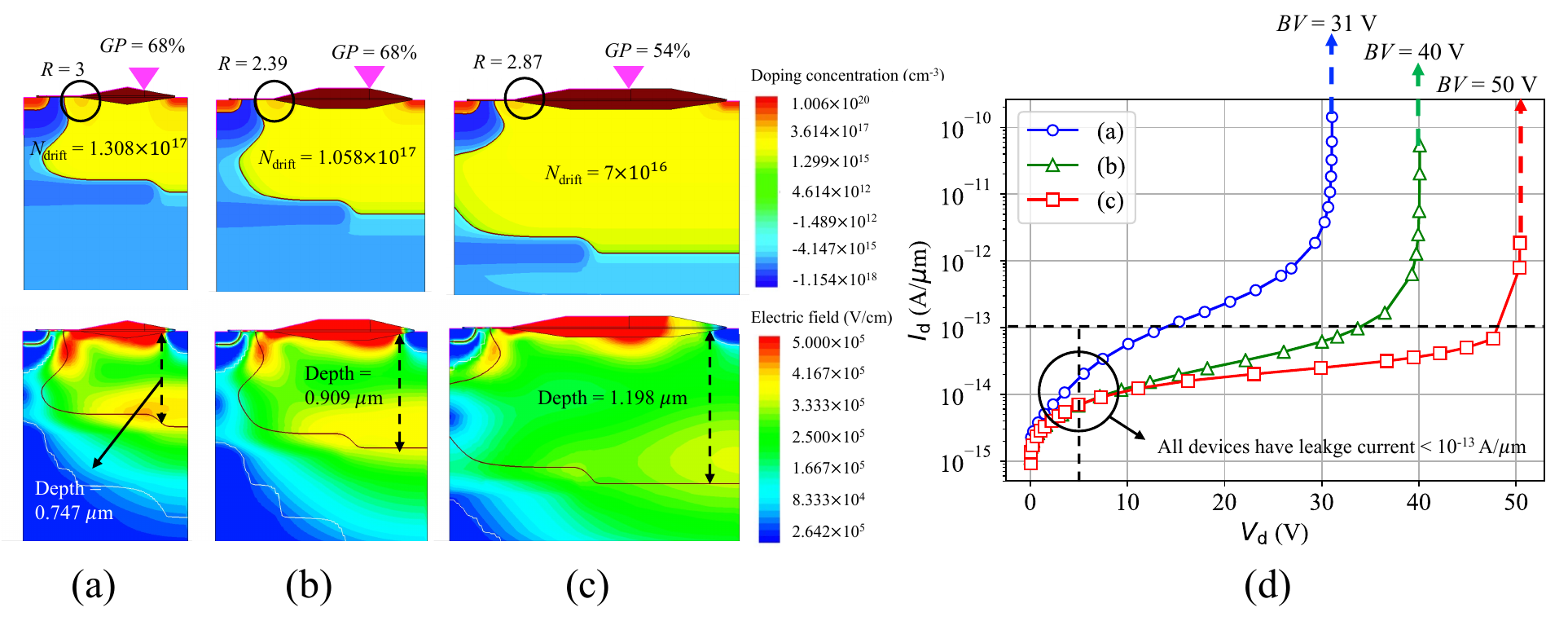}
\caption{\label{fig4} The doping profile and the electrical field distribution for optimized device with (a) $BV$ = 31 V from the unconstrained BO framework, (b) $BV$ = 40 V from our novel algorithm, (c) $BV$ = 50 V from our novel algorithm. (d) Breakdown curves for the three devices.}
\end{figure*}

Fig.~\ref{fig4} (a) - (c) elucidate the device operation of the highest FOM device from the unconstrained BO, the constrained BO with $BV = 40~{\rm V}$ and $BV = 50~{\rm V}$ respectively. The figures on the upper row exhibit the doping profile, while the figures on the lower row are electrical field distributions at the breakdown condition. Inspecting the highest FOM device in the unconstrained BO, we find that the BO drives the device structure result on a diamond-shaped LOCOS as illustrated in Fig.~\ref{fig4} (a) compared to the elongated diamond-shapes in Fig.~\ref{fig4} (b) - (c). However, when constraining the $BV$ to 40~V or 50~V, elongated-diamond shape devices are identified as having the highest FOM. Moreover, the results obtained from these algorithms align with the physical intuition of designers. For example, a higher $BV$ device requires a lower drift doping concentration. The BO enables the automatic identification of which overall structure yields the highest FOM, without relying on time-consuming local/manual optimization.

\section{Discussion}

%The exploration of devices with a range of $BV$ is an important task in power device design. The unconstrained BO yields the highest FOM design in the design space but fails to explore the higher $BV$ design space as shown in Fig.~\ref{fig2} (a). Fortunately, the constrained BO guides TCAD simulations toward a specific $BV$ value as shown in Fig.~\ref{fig2} (b) and (c). The algorithm generates multiple high $BV$ device designs, providing us with valuable information in this high-dimensional design space.

%Fig.~\ref{fig3} shows devices at $BV$ = 30 - 40 V have high FOM $\approx$ 300 $k$W/$\mathrm{mm}^2$. This result indicates the optimized device with $BV$ = 40 V has a comparable FOM with the device found in our unconstrained BO in Fig.~\ref{fig2} (a). This optimized $BV$ = 40~V device design cannot be found by unconstrained BO. The high FOM devices within the range of $BV$ = 30 - 40 V exhibit similar performance and they do not hit any bounds. This indicates that we discovered an extremum region within the 9-dimensional design space. Within this extremum region, we can design devices with $BV$ ranging from 30 V to 40 V and achieve FOM values close to 300 $k$W/$\mathrm{mm}^2$.

Comparing the results in Fig.~\ref{fig2} and \ref{fig3}, we observe that for the 30 V optimized device, the FOM results are almost identical. However, when comparing the 40 V optimized device designs, we can see that the FOM in Fig.~\ref{fig2} (b) is approximately 270 $k$W/$\mathrm{mm}^2$, while it is around 299 $k$W/$\mathrm{mm}^2$ in Fig.~\ref{fig3}. The higher FOM obtained in Fig.~\ref{fig3} is not entirely surprising because 800 iterations were used in Fig.~\ref{fig3} compared to the 200 for Fig.~\ref{fig2}. Seemingly, the 30~${\rm V}$ BV region is easier to access for the BO compared to the 40~${\rm V}$ region. Nevertheless, from our study it appears that an informed constrained optimization can aid BO.

By adding a Lagrange multiplier to the BO, we have successfully been able to perform constrained BO. Specifically, using TCAD simulations, we have effectively been able to constrain the $BV$. Our problem is a specific example of a constraint that requires resource-intensive measurements or evaluations. A previous approach to constrained BO added an additional surrogate function to approximate the output~\cite{gardner2014bayesian}, but in this approach, as the number of constrained output variables increases, more surrogate functions are needed to approximate each output individually. Our constrained BO algorithm is able to solve the specific design problem without any additional computational cost. 

%In this study, a output characteristics, which is also hard to evaluate, is constrained in the optimization problem. We obtain the optimized result with a specific output constraint by adding an Lagrange multiplier into the objective function. We adapt this novel algorithm on the LDMOS design with a target $BV$, which is a very common scenario in industry and it is a time-consuming process to optimize the device with the output constraint. This algorithm not only can be applied to the design of other electronic devices with similar constraints but also can be adapted on the process optimization with a similar limitations. It offers an alternative perspective for solving traditional optimization problems with constraints, opening up new avenues of thinking. And it incredibly accelerates the development of LDMOS design toward the physical limitation on the target output characteristics.

\section{Methods}

%We provided the details of TCAD simulation and the settings for BO, followed by defining our design space. Then, we introduced the BO and TCAD framework and its limitations. We introduce Lagrange multipliers. Finally, we presented our novel algorithm that combines BO with Lagrange multipliers to address the limitation on previous work.

\subsection{TCAD simulation}
\subsubsection{Device structure simulation}\label{Device structure simulation}

In Fig.~\ref{fig:device structure} (a) Nine parameters are adjustable as input parameters for our TCAD simulations. 1) the peak doping concentration ($x_1$ = $N_\mathrm{drift.1}$) of the first Gaussian doping, 2) the length of the first Gaussian doping peak line ($x_2$ = $L_\mathrm{drift.1}$), 3) the length of the second Gaussian doping peak line ($x_3$ = $L_\mathrm{drift.2}$), 4) the gate position ($x_4$ = $GP$), 5) the length of the JFET region ($x_5$ = $L_\mathrm{JFET}$), 6) the length of the FOX ($x_6$ = $L_\mathrm{FOX}$), 7) the doping concentration value on the surface of drift region ($x_7$ = $N_\mathrm{surface}$), 8) the thickness of the FOX ($x_8$ = $T_\mathrm{FOX}$), 9) the shape ratio of LOCOS ($x_9$ = $R$). Among these input parameters, $L_\mathrm{drift.1}$, $L_\mathrm{drift.2}$, $GP$, $L_\mathrm{JFET}$, $L_\mathrm{FOX}$ are layout related parameters which can be defined by photo masks. Nevertheless, $N_\mathrm{drift.1}$, $N_\mathrm{surface}$, $T_\mathrm{FOX}$, $R$ are process related input parameters. The $n$-type dose for the first Gaussian doping determines $N_\mathrm{drift.1}$ and $N_\mathrm{surface}$. The local oxidation process determines $T_\mathrm{FOX}$ and $R$. The shape ratio ($R$) is defined to be equal to $L_\mathrm{step}$/$T_\mathrm{FOX}$, where $L_\mathrm{step}$ is the taper length of LOCOS. $R$ is used to describe the shape of the bird's beak which has significant impact on the performance of the LDMOS transistors. The larger $R$ causes a slender and elongated beak, whereas a smaller $R$ yields a shorter and stout beak. Therefore, various shapes of field oxide structures and doping profiles are considered in our simulations.

\begin{table}
\centering

\caption{\label{tab:table1}Symbols, descriptions, and bounds of the nine input parameters for the BO}
\begin{ruledtabular}
\begin{tabular}{p{2cm} p{3.2cm} p{1.8cm}} 
Symbol (units)&Description&Bound\\
\hline
\addlinespace % Adds space between rows
$N_\mathrm{drift.1}$ (cm$^{-3}$) & The peak value of the 1$^\mathrm{st}$ Gaussian doping & 7$\times 10^{16}$- 2.5$\times 10^{17}$\\
$L_\mathrm{drift.1}$ (nm) & The length of the 1$^\mathrm{st}$ Gaussian doping & 250 - 2700\\
$L_\mathrm{drift.2}$ (nm) & The length of the 2$^\mathrm{nd}$ Gaussian doping & 0 - 500\\
$GP$ (\%) & Percentage of the FOX covered by the gate & 10 - 99\\
$L_\mathrm{JFET}$ (nm) & The length of JFET region & 0 - 700\\
$L_\mathrm{FOX}$ (nm) & The length of field oxide (FOX) & 750 - 2000\\
$N_\mathrm{surface}$ (cm$^{-3}$) & The doping concentration value on the surface of the drift region & 1$\times 10^{16}$- 6$\times 10^{16}$\\
$T_\mathrm{FOX}$ (nm) & The thickness of the FOX & 50 - 150\\
\addlinespace % Adds space between rows
$R$ & The tangent of the angle of the FOX & 0.5 - 5\\
\end{tabular}
\end{ruledtabular}
\end{table}

Fig.~\ref{fig:device structure} (b) illustrates the fixed parameters during the optimization. We use a uniform $p$-type with $2\times 10^{16}$ cm\textsuperscript{-3} doping concentration as the substrate. A $p$-type buried layer with a doping concentration of $9\times 10^{16}$ cm\textsuperscript{-3} is introduced. The length of the $p$-buried layer is fixed to $L_{p\mathrm{-buried}}$ = 0.95 $\mu$m. A high $p$-type doping concentration of $2\times 10^{18}$ cm\textsuperscript{-3} is applied on the surface as the $p$-well for the LDMOS with a fixed depth $D_{p\mathrm{-well}}$ = 0.4 $\mu$m and a fixed length $L_{p\mathrm{-well}}$ = 0.1 $\mu$m. The first drift Gaussian doping peak is positioned at a depth of 0.3 $\mu$m, forming the primary drift region. The second $n$-type doping has a doping concentration of $N_\mathrm{drift.2}$ = $5\times 10^{17}~{\rm cm}^{-3}$, creating the channel in cooperation with the $p$-well. The third $n$-type doping is located beneath the drain region, and it shares the same doping concentration as the second $n$-type doping. This arrangement serves to reduce the gradient of the doping concentration between the drain and the primary drift region so that the electric field on the drain side can be decreased. The thin gate oxide thickness is 12 nm. A LOCOS is implemented on top of the main drift region to increase the $BV$ of the LDMOS transistors. The gate is placed on the elongated-diamond, allowing for the application of gate voltage stress on both the $p$-well and the drift region to control the channel in on-state and electric field distribution on the device surface in off-state. The source and drain regions, featuring an $n$-type doping concentration of $2\times 10^{20}~{\rm cm}^{-3}$, are positioned at the two terminals of the device with a same length $L_\mathrm{source}$ = $L_\mathrm{drain}$ = 0.1 $\mu$m. The definition of the half-pitch as the distance between the source and drain. 

We minimize the number of input parameters in order to simplify the complexity of the TCAD simulation. Some parameters are dependent on defined input parameters. The depth of the drift region ($D_\mathrm{drift}$) is determined by the doping concentration value on the surface of the drift region ($N_\mathrm{surface}$) and the peak value of the first Gaussian doping ($N_\mathrm{drift.1}$) since the first doping profile is a Gaussian distribution. The depth of the drift region is equaling to
\begin{equation}
    D_\mathrm{drift} = 0.3\mu{\rm m} + 3\sqrt{\frac{0.045}{\ln{(N_\mathrm{drift.1}/N_\mathrm{surface})}}} \mu{\rm m}.
\end{equation}
The thickness of the LOCOS ($T_\mathrm{FOX}$) and the shape of tangent of the angle of the FOX ($R$) determines the taper length of LOCOS ($L_\mathrm{step}$) in Fig.\ref{fig:device structure} (a) through $L_\mathrm{step} = T_\mathrm{FOX} \times R$.

\subsubsection{TCAD Physical Models}

For our simulations, we utilize a commercial drift-diffusion software with default silicon parameters\cite{Sentaurus}. To accurately capture various physical mechanisms, we use generation-recombination model, including the doping-dependent and the temperature-dependent Shockley-Read-Hall model, and the Auger model, each addressing specific aspects of carrier behavior and recombination processes. The van Overstraeten model\cite{van1970measurement} is used to account for the impact ionization process. To account for mobility degradation at the silicon-insulator interface, we employ the Lombardi model. In the bulk region, we utilize the widely accepted Philips unified mobility model, which provides a comprehensive description of carrier mobility. Carrier distributions are modeled using the Fermi-Dirac distribution, including the bandgap narrowing model.

\subsubsection{Device Characteristics}
We define the breakdown voltage ($BV(\bm{x})$) and the specific on-resistance ($R_{\rm sp(on)}(\bm{x})$). $BV$ represents the maximum voltage between the source and drain that the device can sustain without experiencing avalanche breakdown when it is in the off-state. We perform the $I_\mathrm{d}$-$V_\mathrm{d}$ measurement and ramp up the $V_\mathrm{DS}$ to extract the maximum $V_\mathrm{DS}$  when the TCAD no longer converges:
\begin{equation}
    BV(\bm{x}) = \mathrm{max}(V_\mathrm{DS}),\quad \mathrm{when} \quad V_\mathrm{GS}=V_\mathrm{BS}=V_\mathrm{SS}=0 ~\mathrm{V}.
\label{eq4}
\end{equation}
We calculate the specific on-state resistance ($R_\mathrm{sp(on)}(\bm{x})$) by 
\begin{equation}
    \begin{aligned}
        R_\mathrm{sp(on)}(\bm{x}) = R_\mathrm{ds(on)}(\bm{x}) \times {\rm AREA}  
    \end{aligned}
\end{equation}
where $AREA$ is half-pitch (HP) times a unit length (default value is 1 $\mu$m on the $z$-direction since we are performing 2D simulations) and $R_\mathrm{ds(on)}(\bm{x}) = V_\mathrm{DS}(\bm{x})/I_\mathrm{DS}(\bm{x}), \quad \mathrm{when} \quad V_\mathrm{GS}=5~\mathrm{V} \quad\mathrm{and}\quad V_\mathrm{DS}=0.1~\mathrm{V}$. We defined the current between source and drain as the leakage current $\quad \mathrm{when} \quad V_\mathrm{GS}=0~\mathrm{V} \quad\mathrm{and}\quad V_\mathrm{DS}=5~\mathrm{V}$. The leakage current is defined at $V_\mathrm{DS}=5~\mathrm{V}$ in order to simplify the simulation process and shorten the simulation time. We fixed the $p$-well doping profile and the second drift doping so that the channel length and the leakage current is fixed in all of our devices. All of our devices exhibit a leakage current less than $10^{-13}$ $\mathrm{A}$/$\mu$m. Leakage current is not the focus of our present study and could be further improved by adjusting the doping concentration of $p$-well and the second Gaussian doping in the drift region.

Lastly, the FOM is the most important metric in this study, equaling 
\begin{equation}
    \mathrm{FOM}(\bm{x}) = \frac{BV(\bm{x})^{2}}{R_\mathrm{sp(on)}(\bm{x})}.
\end{equation}

\subsection{Bayesian Optimization}

%BO is a sequential optimization algorithm for finding the global optimal point of a black-box objective function which is expensive to evaluate. TABLE.~\ref{tab:table2} lists a standard procedure of a sequential optimization. Starting with an empty dataset, the BO loop is repeating three actions: adapt a policy to select the next sampling point, observe the result on the chosen point, and update the dataset. The Expected Improvement (EI) is utilized as the acquisition function to determine the next data point. Subsequently, a real experiment is conducted to obtain the observed result which are then used to update the dataset by Bayes theorem. We use the Gaussian Process Regression ($GPR$) as our surrogate model. Finally, if the stopping conditions are met, such as reaching the stopping criteria, maximum iterations or exhausting the budget, BO will report the maximum result and its corresponding input parameters in the design space. 

For details of BO, we refer to~\cite{garnett_bayesoptbook_2023, frazier2018tutorial}. We utilize the open-source package, \textit{skopt}. We use Gaussian Process Regression (GPR) as the surrogate model and the Expected Improvement (EI) as the acquisition function. The radial basis function (RBF) kernel with length 1.0 is used in the GPR. For the acquisition function, the Limited-memory Broyden–Fletcher–Goldfarb–Shanno (L-BFGS) algorithm is used and 20 iterations are executed to find the extremum of the acquisition function.%, to implement BO, allowing us to minimize the amount of code modification required for BO implementation.

\subsection{Lagrange Multiplier}
%The Lagrange multiplier technique introduces additional scalar variables to transform a constrained optimization problem into another un-constrained optimization problem \cite{bertsekas2014constrained}. This concept is widely used in statistical thermodynamics\cite{reif1967fundamentals}. The newly introduced scalar variables are referred to as Lagrange multipliers. For example, in Fig.~\ref{fig:LM}, assuming we are solving a optimization problem in 2-D. We want to obtain the maximum of a function $f(\bm{x})$, where $\bm{x}$=($x_\mathrm{1}$, $x_\mathrm{2}$) while a constraint called $g$($\bm{x}$) is equal to a constant $c$. This question can be described as 

%\begin{equation}
%    \bm{x^{*}} =  \argmax_{x\in \bm{\R^{2}}} f(\bm{x}) \\
%    \quad \mathrm{subject\quad to} \quad g(\bm{x}) = c
%    \label{eq7}
%\end{equation}

%We assume $f(\bm{x})$ is a convex function so we can calculate the extremum by solving $\nabla	f = 0$. However, we have a constrained function $g(\bm{x})$ which should be a constant value on the point. Therefore, we introduce a function called "Lagrangain ($\mathscr{L}$)", which can be written as 

\begin{figure}
\includegraphics[width=8cm]{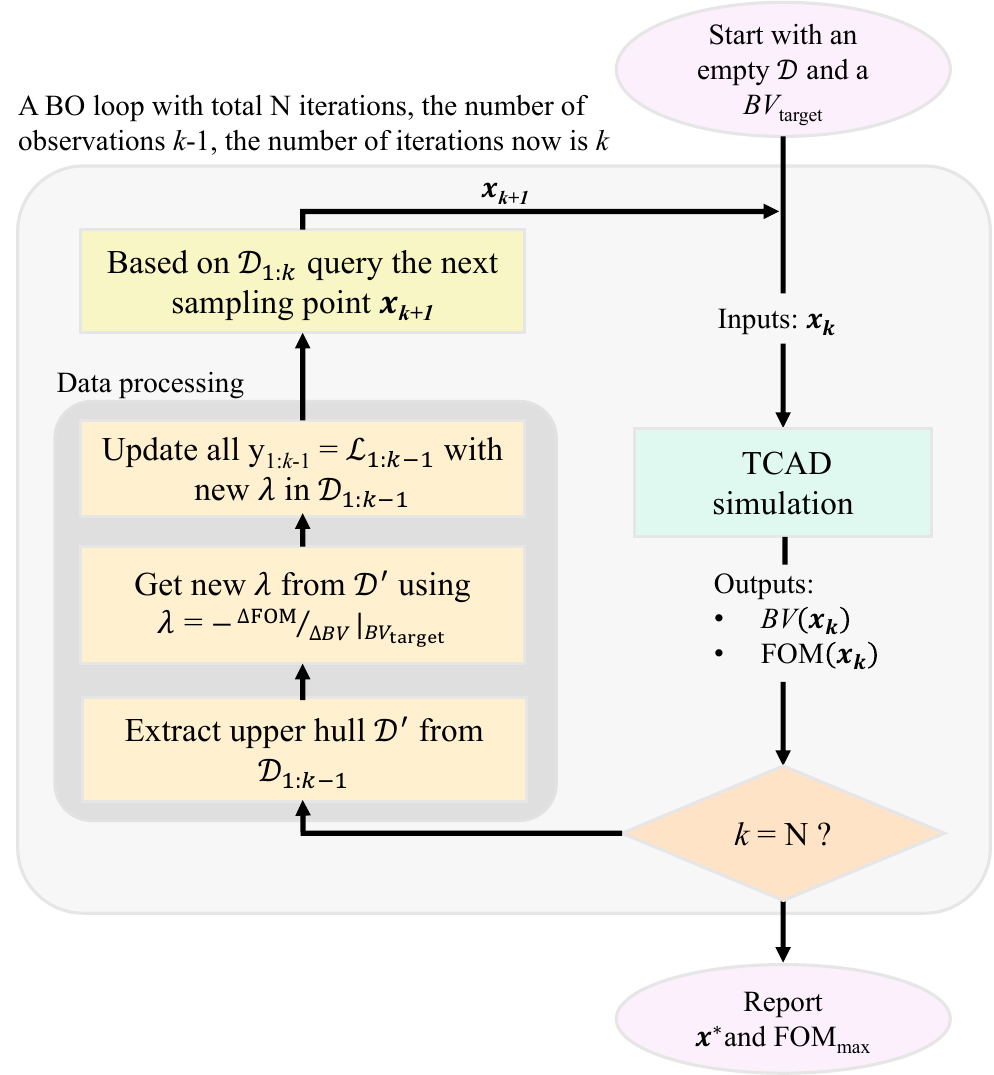}
\caption{\label{fig:Flow} The optimization flow for constrained optimization algorithm. The Lagrange multiplier for the next iteration is determined from the dataset of simulated devices. The $\lambda$ for the first two iterations is set to 0.}
\end{figure}

The algorithm flow chart for the constrained BO problem is shown in Fig.~\ref{fig:Flow}.
The objective function of the BO is the Lagrangian 
\begin{equation}
    \mathscr{L}(\bm{x}, \lambda) =  f(\bm{x}) + \lambda g(\bm{x}) = \mathrm{FOM}(\bm{x})+\lambda (BV-BV_{\rm target})
    \label{eq8}
\end{equation}
where $\lambda$ is the Lagrange multiplier. In our specific problem $f(\bm{x})={\rm FOM}(\bm{x})$ and $g(\bm{x})=BV-BV_{\rm target}$. The Lagrangian function reaches its optimum when $g(x)=0$. Note that when optimizing, $BV_{\rm target}$ is a constant and once $\lambda$ is determined, the last term $-\lambda BV_{\rm target}$ in Eq.~(\ref{eq8}) is just a constant which can be omitted without affecting the results. Correctly determining the Lagrange multiplier $\lambda$ is critical. %Instead of solving $\nabla f = 0$ directly, we solve $\nabla \mathscr{L} = 0$ so that Eq.~\ref{eq8} becomes

We compute the Lagrange multiplier from the simulated data. We determine the devices that form the upper part of the convex hull that contains the FOM vs $BV$ data, yielding a dataset $\mathscr{D}^{'} = \{(BV_{i},\mathrm{FOM}_{i})\}$, where $i$ is an index iterating over the data in the upper hull. The upper hull is illustrated in Fig.~\ref{fig3} where the upper part of the hull consists of a set of 5 points ($i = 1 - 5$). We then identify in which segment of the upper hull the target $BV$ is located, {\it i.e.} $BV_j<BV_{\rm target}<BV_{j+1}$, where $j$ is the number of the segments in the upper hull curve. Finally, the Lagrange multiplier is determined as
\begin{equation}
    \lambda = \left.-\frac{\Delta f}{\Delta g}\right|_{BV_{\rm target}} =-\frac{{\rm FOM}_{j+1}-{\rm FOM}_j}{BV_{j+1}-BV_j}.
\end{equation}

\begin{acknowledgments}
The author also appreciates all of valuable discussion with Dr. Sujatha Sampath of Texas Instruments.
\end{acknowledgments}

\section*{Data Availability Statement}

The data that support the findings of this study are available on request from the corresponding author.

\nocite{*}
\bibliography{aipsamp}% Produces the bibliography via BibTeX.

\end{document}